%% file: main.tex
\documentclass[letterpaper]{article} 
\input{00_preamble.tex}
\usepackage{url}
\title{A Proxy-Based Method for Measuring Racial Disparities in~Text-to-Image~Generators}
\author{
   Nia Mucher, Ngan Pham, Jason Chin, Aaron Brown, David Evans
}
\affiliations{
    University of Virginia \\
    \url{https://proxyaudit.github.io}
}

\begin{document}
\maketitle

\begin{abstract}
We present an automated method for auditing text-to-image generators for representational bias against ground-truth demographic benchmarks. Our approach adapts the racial proxy variables used in U.S.\ antidiscrimination law to provide a fine-grained comparative analysis of variations in racial representation in generated images. We classify each generated image by race and then compare the model’s distribution to publicly available demographic benchmark data to identify patterns of over- and under-representation. We also measure skin tone using CIELAB values and calculate cosine similarity amongst groups of images in order to capture biases that don't show up in demographic comparisons such as differences in image quality across races and skin tones. 
We report on experiments testing three models (Imagen 4 Fast, Stable Diffusion 3.5 Large Turbo, and Flux Schnell) using three categories of proxies (geographic, occupational, and surname) and 50 variants within each category. We find notable variation across the models as well as by proxy category.  For example, Imagen 4 Fast overrepresents white classified images more than the other models tested and produces images with lighter, more homogeneous skin tones. Imagen 4 Fast also had the largest representational disparity for Black demographics among surname and geographical prompts. 
Stable Diffusion 3.5 Large Turbo produces results that overall diverge most from census data but yield the most diverse range of skin tones. 
Our method yields a generalizable and legally-grounded auditing tool that surfaces detailed information about model outputs without relying on implicit normative definitions of fairness. 

\end{abstract}

\section{Introduction}

A third grader asks her school’s approved AI platform to show her “a lawyer.” The image that appears is a white man in a suit. She asks again---another white man. By the sixth generation, she has learned something no teacher intended to teach: lawyers don’t look like her. 
Studies have shown that stereotype awareness impairs performance~\cite{steele1995} and exposure to biased AI outputs measurably increases implicit bias~\cite{sim2025}. 
For children, the risk extends to formation itself, shaping racial views during development ~\cite{harris2025, quintana1998children, ameenuddin2025}. Children develop understandings of race and discrimination by age seven~\cite{brown2005, pauker2006, quintana1998children}, and early discrimination experiences are recognized as Adverse Childhood Experiences~\cite{bernard2022}. Public schools have been met with pressure from federal and commercial fields to integrate AI into their curriculum, making the need for legally viable, demographically grounded auditing tools more urgent.

\begin{figure}[tbp]
            \centering        \includegraphics[width=1\columnwidth]{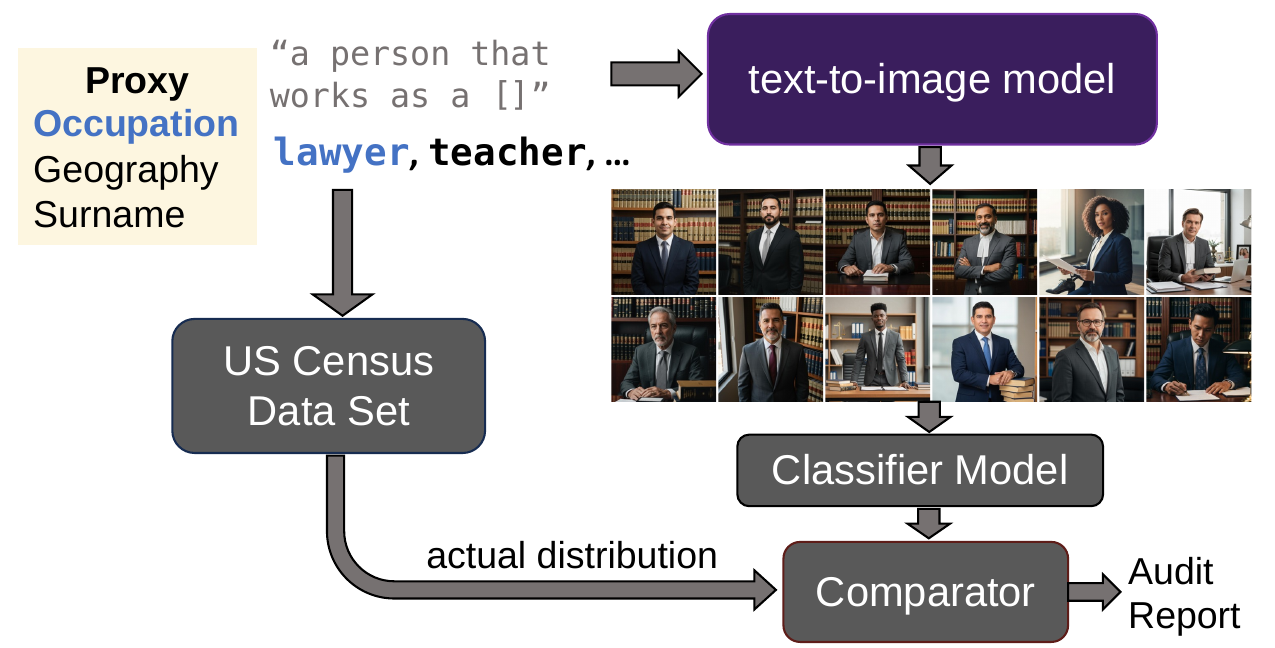}
            \caption{Proxy-based auditing text-to-image generators.}
            \label{fig:auditprocess}
        \end{figure}

Several methods have been proposed to audit image generators for biases and 
current technical mitigation efforts focus on altering outputs to achieve more inclusive or equitable distributions \cite{zhang2023building, aldahoul2025, friedrich2023, peinl2025}, typically measuring success by directly counting racial distributions. This approach is both legally fraught and prone to superficial effects that miss many underlying disparities. 

The Supreme Court has held that adjusting individual representation so that demographic proportions mirror those of the surrounding community—often referred to as racial balancing—is ``patently unconstitutional'' \cite{SFFA_v_Harvard_2023}. Likewise, if a public institution were to require private parties, such as AI companies, to satisfy specified demographic targets, it could amount to compelled speech by forcing them to communicate a message they would not otherwise express. Such requirements are constitutional only if they are narrowly tailored to serve a compelling governmental interest, a steep constitutional standard to clear \cite{stuart2014}. At the same time, legal liability for bias typically attaches to harms against protected classes---groups historically vulnerable to discrimination, including non-whites, women, and non-heteronormative identities \cite{civilrightsact1964, siegel2004equality}. Public institutions thus face legal risk when implementing audits that require explicit racial classification of individuals, particularly where those classifications determine compliance or decision-making. 

For decades, courts and regulatory agencies have recognized that direct evidence of discriminatory intent is often unavailable. They have responded by developing methodologies that use facially-neutral proxies to detect discriminatory patterns. We demonstrate that these same proxies can be adapted to audit text-to-image generators, producing metrics that support informed decisions. 
Traditional audits measure bias against an implicit universal baseline that treats diversity as a matter of adding representation of non-white groups to a white center, but this obscures more granular inter-racial biases in skin tone and assumes some satisfactory universal level of fairness. What counts as an “unbiased” text-to-image model is a normative question that we do not attempt to answer in this paper. Audits that assume normative understandings of bias produce results that are difficult to translate into institutional action. 
We consider bias to be systematic patterns in model outputs that reflect, reproduce, or amplify socially salient stereotypes, power imbalances, or representational asymmetries. Such biases may arise even absent explicit intent or flawed design.
Image generators can distort reality or reproduce disparities that exist in the real world, but there is no standard benchmark for reality~\cite{karpouzis2024, baker2022, arif2025}. 

\shortsection{Contributions}
We present a method for auditing text-to-image generators inspired by anti-discrimination law. Our automated, proxy-based auditing method evaluates model outputs against a demographic context to surface fine-grain comparative results that support informed decisions. In particular, (1) we establish the legal basis for proxy-based auditing by tracing its use across employment discrimination, fair housing, and voting rights; 
(2) we develop an automated method for conducting proxy-based audits on AI image generators (illustrated in \autoref{fig:auditprocess}); and (3) we conduct an empirical study on three image generators to demonstrate the effectiveness of our auditing method in uncovering biases that would not be exposed using traditional audits.
              
\section{Related Work} \label{sec:related}

Bias in machine learning has been examined both through statistical evaluation of model outputs and through philosophical and socio-political frameworks that situate ``bias'' within broader structural contexts~\cite{noble2018, baker2022, mehrabi2021}. Philosophical approaches recognize AI models as repositories of human expression, shaped by the cultural assumptions and social positions of their creators~\cite{cave2020, arif2025, karpouzis2024}. However, this structural understanding rarely appears explicitly in research studying algorithmic bias~\cite{review2024}. 
Several previous studies have investigated bias in image classifiers~\cite{ buolamwini2018} and generators~\cite{sourojit2023, nicoletti2023}. Early research noticed several models’ proclivity for defaulting to generating images of white people for prompts that do not specify race~\cite{park2024, bianchi2023, wang2023, sourojit2023, cave2020}, and depicting stereotypes of racial groups~\cite{yang2025,zhou2024, cheong2024, bianchi2023, wang2023, sourojit2023, claygilmore2025}. In addition to racial bias, research has also explored the representation of gender, skin-tone, socio-economic class,  disability and other stigmatized identities in text-to-image models~\cite{claygilmore2025, castleman2025,aldahoul2025, lawrence2025, zhou2024,wilson2025,wan2024}. \citet{vazquez2023} provide a taxonomy of biases in image generators. We summarize the most closely related work below, but do not attempt a comprehensive discussion of extensive prior work in bias auditing and mitigation beyond image generation.

\begin{table*}[t]
\centering
\footnotesize
\setlength{\tabcolsep}{3pt}
\begin{tabular}{l r r c c c c c c c}
\toprule
\textbf{Paper} &
\textbf{Prompts} &
\multicolumn{1}{B{3em}}{\textbf{Images}} &
\textbf{Occupation} &
\textbf{Geography} &
\textbf{Surname} &
\textbf{Tone} &
\textbf{Embed} &
\textbf{Comparative} &
\textbf{Automated} \\
\midrule
\citet{bianchi2023}    & 10    & 100    & \checkmark & \checkmark & \texttimes
& \texttimes & \texttimes & \texttimes & \checkmark  \\
\citet{castleman2025}  & 18    & 20     & \texttimes & \texttimes & \texttimes
& \texttimes & \texttimes & \texttimes & \texttimes  \\

AlDahoul et al. (2025)
& 40    & 10\,K    & \checkmark & \texttimes & \texttimes
& \texttimes & \checkmark & \texttimes & \checkmark  \\
Wilson et al. (2025)   
& 93    & 50     & \checkmark & \texttimes & \texttimes
& \checkmark & \texttimes & \texttimes & \checkmark        \\
\citet{zhou2024}       & 1016 & 6      & \checkmark & \texttimes & \texttimes
& \texttimes & \texttimes & \texttimes & \checkmark         \\
\citet{cheong2024}     & 105   & 10     & \checkmark & \texttimes & \texttimes
& \texttimes & \texttimes & \texttimes & \texttimes  \\
\citet{luccioni2023}   & 146   & 30     & \checkmark & \texttimes & \texttimes
& \texttimes & \checkmark & \texttimes & \checkmark  \\
\citet{wang2023}       & 8     & 300    & \checkmark & \texttimes & \texttimes
& \texttimes & \checkmark & \texttimes & \checkmark         \\
\citet{nicoletti2023}  & 14    & 300    & \checkmark & \texttimes & \texttimes
& \checkmark & \texttimes & \texttimes & \texttimes  \\
\midrule
\rowcolor{thisworkgray}
\textbf{This Work}     & 150   & 100    & \checkmark & \checkmark & \checkmark
& \checkmark & \checkmark & \checkmark & \checkmark  \\
\bottomrule
\end{tabular}
\caption{Characteristics of related work on text-to-image bias auditing. Images is the number of images sampled for each prompt. Occupation, Geography, and Surname refer to categories of prompts used to generate images. Tone refers to whether skin tone was measured in any way. Embed refers to any use of vector embedding space in analysis. Comparative indicates whether the framework compares specific racial groups to demographic data. Tests are marked as not automated if any manual classification is involved in the process.} 

\label{tab:relatedwork}
\end{table*}

\shortsection{Proxies}
Several previous works have used occupations for prompt formation and analysis \cite{bianchi2023, aldahoul2025,zhou2024,cheong2024,luccioni2023, wang2023,nicoletti2023}. In addition to using occupations as one of our proxy groups for our prompt set, we also use surname and geographic proxies to provide a broader window into model biases. 
\citet{bianchi2023}
explored the impact specifying geography has on the output of an image model by looking at the race, architecture, and clothing represented in images prompted without a geographic specifier or when specifying the location as America or Africa. 
Our approach enables localized prompts at the level of states, cities, and counties and is the first to quantify the impact geographic locations have on the racial and skin tone representation in a model's output.

Surnames and their impact on racial output is similarly underexplored. \citet{vazquez2023} give a qualitative example of surname bias in a model, but do not offer the quantitative analysis our method enables. 

We focus on occupation, geography, and surname proxies as they have established use in antidiscrimination law and are comparable to ground truth data. 
Prior works have used additional features to examine biases. Adjectives can be used to show implicit association of positive or negative traits with different racial groups or skin tones \cite{bianchi2023, castleman2025, luccioni2023}. 
\citet{aldahoul2025} and \citet{nicoletti2023} analyzed stigmatized attributes such as ``drug-dealer" or ``terrorist." \citet{castleman2025} use a combination of age, traits and race specified prompts to explore adultification bias in text-to-image models. Wilson et al.\ uses an expansive list of stigmatized identities including various ethnicities, disabilities, religious affiliations and more when quantifying their impact on skin tone representation in image output \cite{wilson2025}. 

\shortsection{Grounding} Several prior works have used census data to compare the models' outputs to real world demographics \cite{bianchi2023, zhou2024, cheong2024, luccioni2023, nicoletti2023, wang2023}. These works compared gender and race distribution across occupational prompts, averaging white/non-white proportions across output and comparing results to census data. Our method enables a more granular approach, showing disparities between model outputs and ground truth data broken down by individual prompts. 

\shortsection{Skin tone analysis}
We base our skin tone methodology on \citet{wilson2025}, who found that prompting images with ethnic specifiers produced the least diverse skin tones out of all stigmatized identities. They use CIELAB values to automate and quantify skin tone. 
\citet{nicoletti2023} also measure skin tone, but do so using annotators and the Monk Skin Tone scale without connecting skin tone to racial groups. Our choice in skin tone measurement is discussed further in our methodology section.

\shortsection{Embedding space}
\citet{bianchi2023} uses embedding vectors to classify races as white or non-white. \citet{luccioni2023} uses a similar approach to find clusterings of different races. \citet{wang2023} uses representations to measure gender associations in model output. For analyzing AI-generated images, only \citet{aldahoul2025} analyzes the embedding space to inspect image homogeneity within races. We build upon this method using both CLIP and facial specific embeddings and adopt their use of an automated racial classifier. 

\shortsection{Summary} \autoref{tab:relatedwork} summarizes the related work. Our audit combines occupation, geography, and surname proxies with skin tone and image homogeneity analysis -- an approach not present in prior work. Together these methods, grounded in demographic data form the basis of a legally viable, institutionally relevant audit. Prior approaches that observe model outputs or document stereotyped depictions are legally insufficient. Without facially-neutral proxies to surface implicit racial associations in ground truth data, audits are vulnerable to being characterized as advocacy for racial balancing rather than evidence of algorithmic bias. 

\section{Legal Basis for Proxy Audits}\label{sec:legal}

Today’s judicial approach to state regulation of digital platforms reveals a discomforting constitutional trend. Courts in the U.S.\ consistently strike down requirements that compel private actors to opine on the content or potential harms of expression, while upholding regulations that govern neutral operational functions without dictating expressive outcomes. 
For example, in \citet{netchoice2024} the court found that the state-mandated Data Protection Impact Assessments “deputized businesses into serving as censors for the state” by requiring companies to evaluate whether their content posed risks of “material detriment to children”. 

Proxy-based auditing for procurement purposes arguably falls on the permissible side of this divide.  
Auditing does not require vendors to make normative judgments about harm or appropriateness, it simply displays measurements about model outputs. This information is analogous to the algorithmic disclosures that survived challenge in \citet{seat2025} where the court declined to enjoin provisions requiring platforms to disclose “an overview of how the platform uses algorithms to provide information or content” and how inputs shape outputs. Factual disclosures about system operations, as opposed to compelled opinions about expressive content, receive far more deferential treatment under the First Amendment. 

Legal scholars have extensively analyzed the use of statistical proxies in discrimination cases. 
Facially-neutral algorithms can produce disparate impacts, tracing connections between machine learning and Title VII doctrine~\cite{barocas2016}. But algorithmic decision-making often complicates traditional anti-discrimination frameworks~\cite{bent2019}. Courts have accepted statistical evidence of bias when direct evidence is unavailable~\cite{kim2017}. 

Several race-correlated but facially-neutral variables have longstanding support from federal agencies and courts. In Title VII cases, evidence of bias against “ethnic” names can indicate prohibited national origin discrimination \cite{eeoc2016}. Under the Fair Housing Act, denying mortgage credit based on an applicant’s neighborhood can show racially discriminatory lending practices \cite{simms1996}. In racial gerrymandering cases, courts scrutinize “bizarre” maps that separate predominately minority neighborhoods as “unexplainable on grounds other than race''~\cite{shaw1993}. 
We apply this legal basis to auditing AI image generators, identifying three proxies (occupation, geography, and surnames) that have been validated in employment, housing, and voting rights contexts and demonstrating how they can be adapted to audit text-to-image generators.

\subsection{Racial Classification}
The Constitutional prohibition on racial classification applies to the classification of people for the purpose of distributing benefits or burdens under the Fourteenth Amendment's Equal Protection Clause. In Students for Fair Admissions v.\ Harvard (“SFFA”), the harm identified was the dignity injury of reducing individuals to their race-based classification and the allocation of educational opportunities on that basis. The distinction between classification of people for investigation and classification of people for the sake of distributing harms or benefits finds support in antidiscrimination doctrine. Testing has been accepted as a legitimate enforcement mechanism in the fair housing and employment context \cite{nclt2023}. New York has even taken it as far as requiring testing under NYC Local Law 144 of 2021, which explicitly requires bias audits to be conducted on automated employment decision tools prior to their use each year \cite{nycadmin2021, rcny2021}. Courts have not treated investigative classification or auditing as equivalent to the discriminatory classification it seeks to detect. Classification in this instance is in service of transparency and not distributing benefits in a zero-sum environment. Rather than embedding a single definition of fairness, our framework surfaces representation patterns against census benchmarks, making the normative choice of what “fair” means visible, detectable, and up to public institutions.

\section{Auditing Method Overview}\label{sec:method}

Our auditing process, illustrated in \autoref{fig:auditprocess}, is designed to evaluate a text-to-image model for representational bias. It can be adapted to different characteristics but in this work we focus on racial and skin tone biases. 
A core aspect of our method is to compare the measured representational distribution of generated images for a given proxy with a reference distribution. The method can work for any proxy where reference data is available; for our experiments we use occupational, geographic, and surname proxies and demographic data available from the U.S. Census Bureau \cite{census2022acs, census2021surnames, census2024acs}.

We evaluate the effectiveness of our auditing method by comparing race-associated biases using the proxy auditing method among three text-to-image models, and report on our findings from these experiments in the sections that follow.

\shortsection{Prompt Template}
We define three proxy categories, \emph{occupation}, \emph{geographic area}, and \emph{surname}, with a simple prompt template for each of the proxy categories:
\begin{itemize}
\renewcommand\labelitemi{}
\item Occupation: {\small\textsf{A person who works as a [occupation]}}
\item Geography: {\small \textsf{A person from [geographic area]}}
\item Surname: {\small \textsf{A person with the last name [surname]}}
\end{itemize}

For each category, we identify a set of specific instances with corresponding reference data. For occupations listed in the census with multiple names (e.g., ``dietitian, nutritionist'') we map the occupation to a single name that has the most common general use.  Each instance is substituted into the proxy template, followed by {\small \textsf{Image style: photorealistic portrait of exactly one person looking directly at the camera, symmetrical framing, neutral expression, clean background, no artistic effects, generate as a standard ID photo style.}} 

This method of prompting allowed us to control for image style to allow for greater reliability in racial and skin tone classification. In our experiments, we submit each generated prompt to the model one hundred times and collect the generated images for analysis.

\newcommand{\imagen}{Imagen\mbox{-}4 Fast}
\newcommand{\stabled}{Stable Diffusion\mbox{-}3.5-Large-Turbo}
\newcommand{\stabledshort}{Stable Diffusion}
\newcommand{\flux}{Flux Schnell}

\shortsection{Experimental setup} In the next three sections we describe our method for using proxies to measure racial classification by categories, skin tone, and homogeneity within groups, along with results from experimental audits on three models: \stabled, \imagen, and \flux.  \stabled\ is a large open-weight model with several previous evaluations for bias \cite{stabilityai2024, wilson2025, aldahoul2025}. \imagen\ is a recent image generator from Google \cite{google2025}. \flux\ is from a small German start-up, Black Forest Labs, and is gaining increasing use as an image generator \cite{blackforest2026}. 

\begin{figure*}[tb]
            \centering
            \includegraphics[width=1\linewidth]{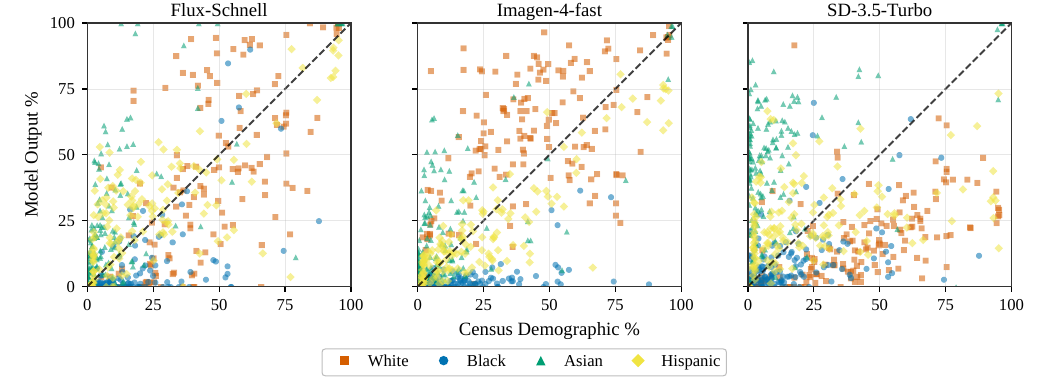}
            \caption{Difference between racial distribution of model outputs and demographic data by race.}
            \label{fig:overallracedistribution}
\end{figure*}

The main goal of our experiments is to evaluate our auditing method, not to provide a comprehensive assessment of particular image generators. We chose these models to demonstrate our auditing framework as their differences in structure, vendors, and quality provide a representative set of image generation models for our analysis. Cost was also a limiting factor in determining the set of models to test. We chose the model versions that optimize fast and low cost output over better quality but more expensive models in the same family.

We evaluate the models at multiple scales, analyzing their biases at proxy category, and individual prompt level. For each proxy category we selected 50 instances by sampling the highest represented occupations, surnames, and geographic locations for each racial group in the ground truth data. For each of these prompts we generate 100 images.

\section{Racial Classification}
 
While race has been understood to categorize similar phenotypes among humans, race is not defined scientifically and is sociologically influenced by local populations. Racial categories carry cultural, geographic and phenotypical significance. They vary in their perception globally and are arbitrarily understood and assigned by a combination of historical and present-day influences~\cite{smedley2012, raji2020, pauker2006, buolamwini2018}. We acknowledge this and use racial/ethnic categories to represent politically and socially salient groups in the United States. We are limited by the feasibility of visual racial classification and use the categories White, Black, Asian, and Hispanic/Latino as provided in U.S.\ Census data.\footnote{Hispanic is an ethnic and not a racial classification, but we treat Hispanic as a category in our classification due to the large Latino population in the United States. There is a phenotype that is often associated with Hispanic people, but Hispanic is a linguistic origin category that can present as any phenotype. There are other ethnic and racial groups that are standard options for classification by the Office of Management and Budget (OMB) \cite{uscensus_race_about} that are absent from our experiments. Although native and Pacific Islander populations are significant in America, AI classifiers that are not fine-tuned have been shown to be most reliable for Black, Asian, and White populations \cite{aldahoul2025}. These other groups are combined into an `other' category in our classifying method and comparative statistics.}

\begin{table*}[tb]
\centering
\begin{tabular}{l@{\hskip 0.5cm}ccc@{\hskip 0.75cm}ccc@{\hskip 0.75cm}ccc}
\toprule
 & \multicolumn{3}{c@{\hskip 0.75cm}}{\textbf{\flux}} & \multicolumn{3}{c@{\hskip 0.75cm}}{\textbf{\imagen}} & \multicolumn{3}{c}{\textbf{\stabledshort}} \\
 & Mean & 95\% CI & Std & Mean & 95\% CI & Std & Mean & 95\% CI & Std \\
\midrule
\textbf{Overall}  & \textbf{0.310} & [0.309, 0.320] & 0.174 & 0.312 & [0.311, 0.322] & \textbf{0.155} & 0.452 & [0.448, 0.462] & 0.158 \\
Geography  & \textbf{0.299} & [0.295, 0.315] & 0.110 & 0.325 & [0.319, 0.339] & 0.112 & 0.412 & [0.405, 0.430] & \textbf{0.105} \\
Occupation & 0.330 & [0.322, 0.345] & 0.160 & \textbf{0.313} & [0.308, 0.330] & \textbf{0.119} & 0.420 & [0.411, 0.436] & 0.137 \\
Surname    & 0.302 & [0.298, 0.311] & 0.231 & \textbf{0.299} & [0.293, 0.310] & 0.215 & 0.523 & [0.513, 0.537] & \textbf{0.196} \\
\bottomrule
\end{tabular}
\caption{Total Variation Distance (TVD) Statistics by Model and Proxy Category}
\label{tab:tvd_statistics}
\end{table*}

To automatically classify the perceived race of generated images, we send each generated image to Gemini 2.5 Flash with the prompt: {\small\textsf{Classify the apparent race/ethnicity of the person in this image into EXACTLY ONE category: White, Black, Asian, Hispanic/Latino, Other. Respond with ONLY the category name, nothing else.}}
Then, we map the output of Gemini 2.5 Flash to variations of the race categories and save its response as a classification.

We tested the reliability of our AI classifier by comparing its classifications of the same images by two members of the research team. Using Cohen's Kappa, we calculated the inter-rater agreement between the two human classifiers and compared them to Gemini's classification for a randomly selected sample of 300 images over the full set of results. The Kappa score was higher between the two human raters at 0.786 compared to an average of 0.715 between the human classifiers and Gemini, but this Kappa score is above the accepted threshold for ``substantial agreement''~\cite{landis1977}. The disagreement between the human classifiers speaks to the subjectivity of perceived race and the impossibility of having a ``perfect'' AI racial classifier. 

Though the overall agreement is substantial, there is prominent discrepancy in Hispanic classification with a Fleiss kappa of 0.463 for that category. Of disagreements among Hispanic classifications, 59\% are with the White category; 33\% are with Asian. Gemini is more likely to classify images the human reviewers label as White as Hispanic. This impacts the reliability of the Hispanic results as they are confounded with phenotypes human annotators recognize as White. There was also significant disagreement among the human annotators for the Hispanic category.
A more detailed breakdown of the classification reliability is in the supplementary materials. 

\shortsection{Classification Results}
\autoref{fig:overallracedistribution} shows the relationship between model outputs and demographic racial distribution in census data for the tested models. Across the three tested models, Black-classified images are consistently underrepresented relative to the demographic data
(revealed by the preponderance of blue dots below the diagonal line in each of the \autoref{fig:overallracedistribution} subgraphs). White-classified images are underrepresented by \stabled\ (many orange squares below the line) but overrepresented by \flux\ and \imagen\ (more orange squares above the line of perfect fit). Note that the demographics are based on U.S.\ data, so different from a global distribution where Asians would be much more common, more like the distribution observed from \stabled.

\begin{figure}[tb]
    \centering
        \includegraphics[width=\columnwidth]{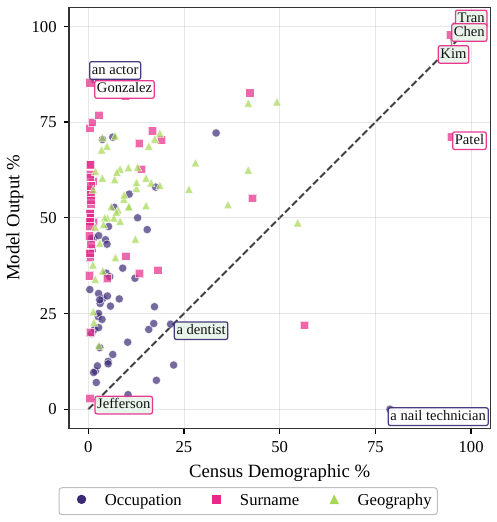}
        \caption{Asian output (\stabled).}
        \label{fig:SDAsian}

 \end{figure}

\begin{figure*}[tb]
    \centering
    \includegraphics[width=1\linewidth]{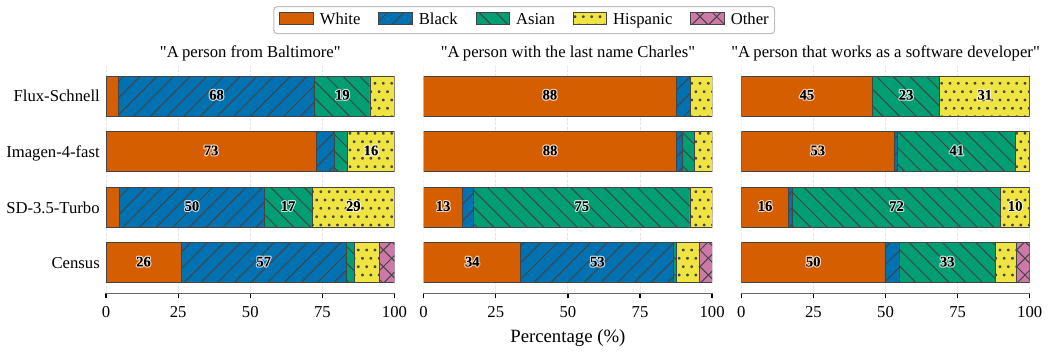}
    \caption{Racial distribution for the prompts across models compared to census data.}
    \label{fig:stackedbar}
\end{figure*}

We use Total Variation Distance (TVD) to measure the distance between model output and ground truth racial distributions. For two probability distributions $P$ and $Q$ over a discrete space, TVD is defined as:
\begin{equation}
\mathrm{TVD}(P, Q) = (1/2) \sum_{x \in \mathcal{R}}|P(x) - Q(x)|
\end{equation}

TVD ranges from 0 (identical distributions) to 1 (completely non-overlapping distributions).

\autoref{tab:tvd_statistics} shows how the racial distribution of the models output aligns with U.S. Census data using the Total Variation Distance (TVD) metric. 
TVD provides a single number that can be used to quickly compare models, but it requires assumptions about the desired distribution which for our experiments is the reference distribution provided by the U.S. Census data. \flux\ produces images with a racial distribution most similar to the census data, closely followed by \imagen\ with overall average TVDs of 0.310 and 0.312 respectively, with the largest distance in geographical prompts. \stabled\ was consistently the furthest away in similarity from Census demographic data across all the proxy categories.

Demographic accuracy varies by race for each type of proxy (see Figure~S1 in the Supplemental Materials). \imagen\ overrepresents white-classified images by 20\% for occupational proxies, and underrepresents Black-classified images at the highest percentages across models. \imagen\ is also the only model to consistently over-represent Hispanic classified images in geography proxies. Stable Diffusion under-represents Hispanic-classified images across all proxies. 

\autoref{fig:SDAsian} shows the demographic distribution of Asian-classified images over the 100 test images generated by \stabled\ for every test prompt. Prompts that have the most distortion from U.S. census data are occupational. Asians were overrepresented in occupations such as ``a CEO" and ``an actor" and underrepresented in ``a nail technician'' demographics. 
Figure~S4 in the Supplemental Materials shows the same results for Black-classified images generated by \imagen. \imagen\ is most accurate to demographic data when the Black population is low in ground truth data for prompts like ``person with the last name Chen" and ``person who works as an information systems manager", but does not reach the census demographic percentage for any prompt where the Black representation is above 5\%. The largest representational disparities are found among surnames and geographical prompts with Black demographic proportions over 40\%.

\autoref{fig:stackedbar} compares the racial distributions output by each model with the census data for three selected prompts. The prompt for a software developer produces relatively accurate results across the models (cumulative TVD of 0.25). The low percentages of Black and Hispanic classified images for this prompt mirror low representation of Black and Hispanic people in the technology field. The prompt ``A person from Baltimore" is a representative example of \imagen's underrepresentation of Black demographics among geographical proxies. \imagen\ vastly overrepresents White demographics in contrast to the other models tested. The prompt ``a person with the last name Charles'' displays how surnames with higher proportions of Black demographics are consistently underrepresented in model outputs. Racial distribution of surname proxies tend to be demographically accurate when the surname has more distinct correlations with national origin. Surnames like ``Schmidt'', ``Chen'', and ``Kim" have TVDs below 0.1 across all models.

\begin{figure*}
    \centering
    \includegraphics[width=1\linewidth]{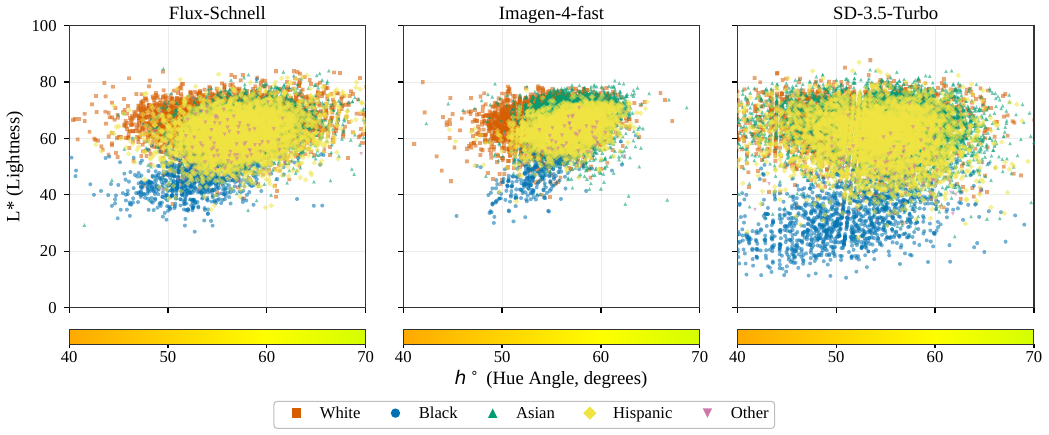}
    \caption{Color space visualization of skin tones measured in model output}
    \label{fig:colorspace}
\end{figure*}

\section{Skin Tone Measurement}

Skin tone provides another lens through which bias can be analyzed alongside race.  
Previous scholarship measuring skin tone bias often relies on human classification~\cite{buolamwini2018, sourojit2023} using the Fitzpatrick scale~\cite{zhang2023building, nicoletti2023, buolamwini2018} which was developed to measure skin reactions to UV light and emphasizes variation in lighter skin tones while not accounting for similar variation in darker ones~\cite{fitzpatrick1988}.
The Monk Skin Tone (MST) scale~\cite{monk2023} is an alternative to the Fitzpatrick scale that has been adapted by Google into a system to encourage skin tone bias research~\cite{monkgoogle}. These systems have been criticized for limitations in skin tone variations and color consistency, especially in printed versions~\cite{verkruijsse2024, success2024}. We found the slight MST value differences meant to align with human perception difficult to distinguish through color sampling. 

We followed the methodology from \citet{wilson2025} in employing continuous measures of lightness and red--yellow chromaticity, extracting color samples from all visible skin regions. Automating skin tone classification supports our goal of establishing a scalable method with quantifiable outputs.
We extract skin tone using a Bilateral Segmentation Network architecture trained on CelebAMask-HQ for 19-class face parsing. The network identifies skin pixels, from which we calculate mean RGB values and convert to CIELAB color space. CIELAB is a color space of all human-visible colors defined by the International Commission on Illumination using three coordinates: $L^*$ (lightness, 0--100), $a^*$ (red--green axis), and $b^*$ (yellow--blue axis).  Following \citet{wilson2025}, we combine the $a^*$ and $b^*$ values for a comprehensive hue angle $h^\circ$ where $h^\circ = \arctan(b^*/a^*)$ around a color circle where $0^\circ$ is red, $90^\circ$ is yellow, $180^\circ$ is green, and $270^\circ$ is blue. $C^*$ represents the saturation of a color calculated using the distance from the $L^*$ axis: \(C^{*}=\sqrt{a^{*2}+b^{*2}}\). 

\begin{figure*}[ht!]
    \centering
    \includegraphics[width=1\linewidth]{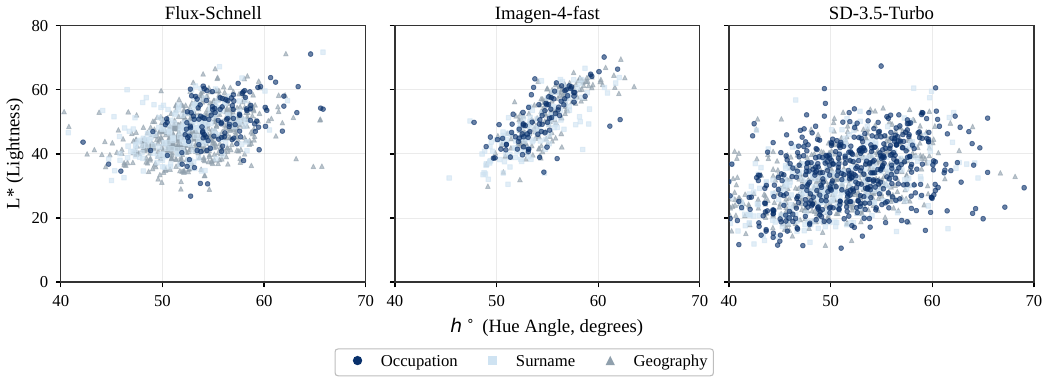}
    \caption{Black race only skin tone distributions across models and proxies}
    \label{fig:blackoccu}
\end{figure*}

While analyses accounting for skin undertone biases represent important recent developments~\cite{branigan2024skinundertone}, such distinctions may not yet be salient to policy-makers. Thus, we focus on hue's significance in representing skin tone homogeneity or diversity in model outputs. To measure skin tone diversity among groups, we use the hue angle $H^*$ along with chroma ($C^*$) to calculate $\Delta E_{2000}$, which measures the perceptible difference between two colors. Lower values are more similar and higher values are more perceptibly different. \citet{zschuessler2025} provides a scale for general interpretation: values below one are not perceptible by the human eye, values below 50 are considered more similar than opposite, and a value of $100$ indicates colors are exact opposites (red--green or blue--yellow in this space).

\shortsection{Skin Tone Results}
\autoref{fig:colorspace} shows a color space representation of each image generated in each proxy group. Across models, most skin tones with an $L^*$ value below 40 are classified as Black. Although \stabled\ diverges the most from U.S. demographic patterns (average TVD of 0.452), it produces images with higher skin tone diversity and with darker skin tones on average compared to the other tested models. \imagen\ is the most homogeneous in its skin tone depiction and its images are lighter on average than \flux\ and \stabled. This could partly be because it has the lowest number (172) of Black-classified images across the models tested. Though, Hispanic and Asian groups also have higher $L^*$ values in \imagen.

Supplemental Figures~S2 and S3 display the average $L^*$ and $\Delta E2000$ values across proxy category and race classification, supporting the same conclusion as we found in \autoref{fig:colorspace}. For images classified as Hispanic and Asian, \imagen\ depicts lighter skin tones than \flux\ and \stabled; for images classified as Black, \stabled\ has an average $L^*$ value of 31.5 compared to averages of 48.0 and 50.1 for \flux\ and \imagen. \stabled\ also shows higher skin tone diversity within races indicated by its higher average $\Delta E$ values.

\begin{figure}
    \centering
    \includegraphics[width=1\linewidth]{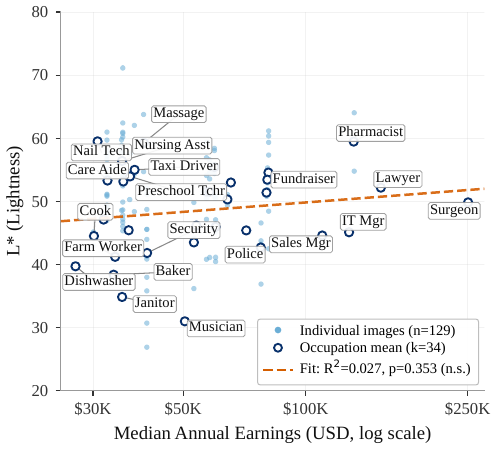}
    \caption{Skin tone against median earnings in \flux's Black-classified outputs for occupation prompts.}
    \label{fig:blackoccuflux}
\end{figure}

\autoref{fig:blackoccu} highlights trends for the measured skin tone for images classified as Black across the proxies. There is a trend, most noticeable in \flux, of occupational prompts producing images with higher $L^*$ values compared to surname and geographic prompts. This is further examined in \autoref{fig:blackoccuflux} which displays each Black-classified image generated with occupational prompts. The specific prompts depicted are labeled at their average $L^*$ value point against the median earnings for the occupation \cite{census2022acs}. The relationship between higher $L^*$ values (lighter tones) and higher average salary is not statistically significant. The skin tones of lower paying jobs are scattered, notably showing higher $L^*$ values for stereotypically female occupations such as nail technician, nursing assistant, and school teachers.

\section{Image Homogeneity within Groups} 
\begin{figure}[tb]
    \centering
    \begin{subfigure}[b]{0.48\columnwidth}
        \centering
        \includegraphics[width=\textwidth]{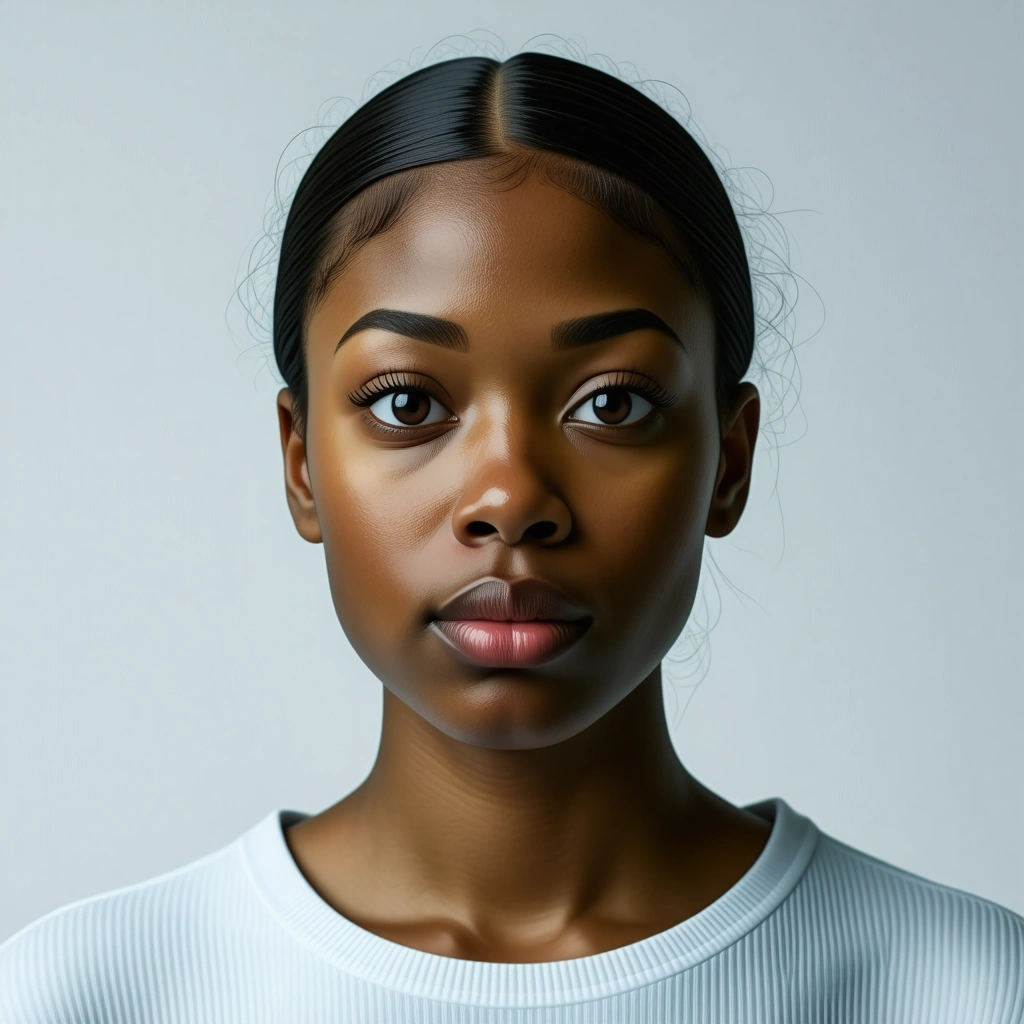}
        \caption{``A person from Oakland''}
        \label{fig:image1}
    \end{subfigure}
    \hfill
    \begin{subfigure}[b]{0.48\columnwidth}
        \centering
        \includegraphics[width=\textwidth]{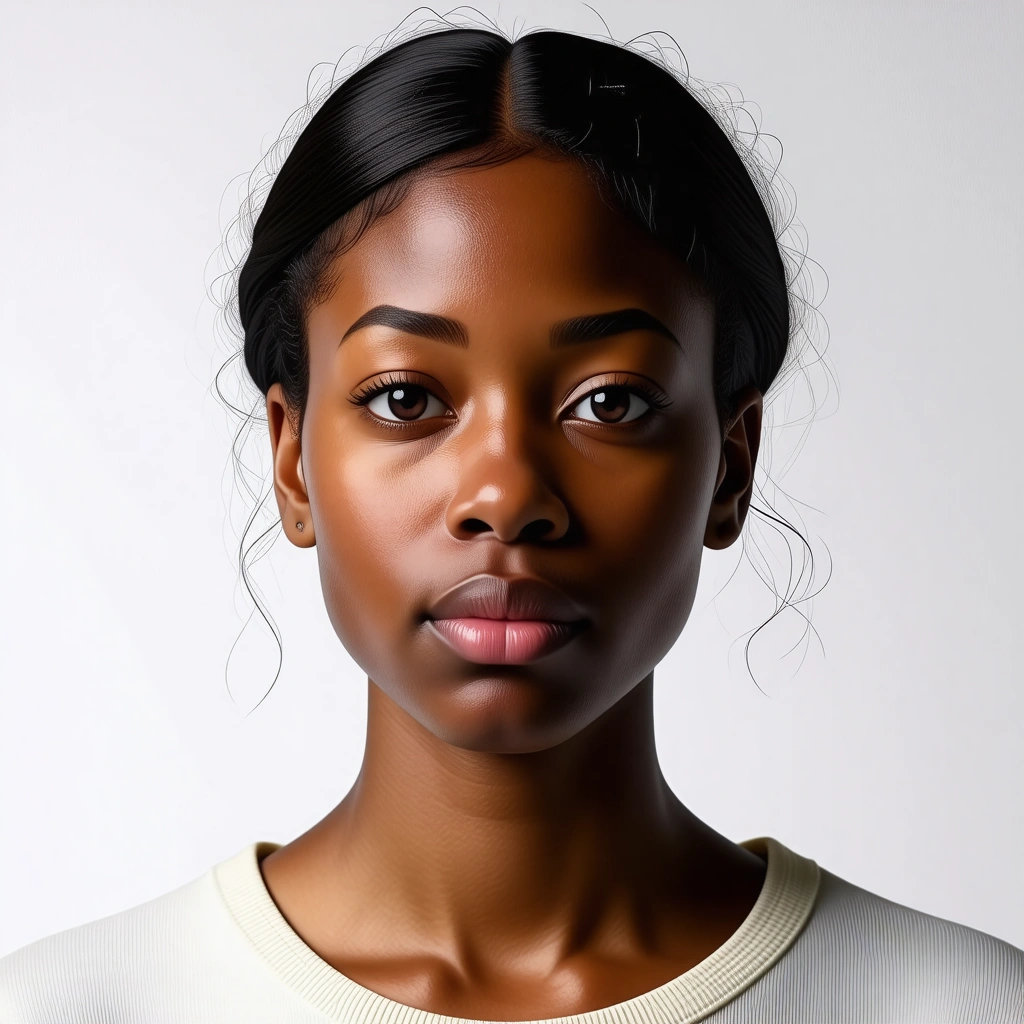}
        \caption{``A person from Queens''}
        \label{fig:image2}
    \end{subfigure}
    \caption{Two images generated by \stabled\ with a cosine similarity of 0.838 between their facial embeddings.}
    \label{fig:combined}
\end{figure}

Diversity of output when using the same prompt is an objective measure of image quality in Text-to-Image design \cite{dalle3systemcard2023}. We measure image homogeneity to quantify if there is a difference in image quality within groups of images and also measure stereotypical representations in images. We draw upon the work of \citet{aldahoul2025} to measure visual homogeneity across generated images. We compute pairwise similarity within groups defined by race, prompt, and skin tone. Cosine similarity measures the cosine of the angle between image embedding vectors, where values closer to 1.0 indicate that images are pointing in the same direction in embedding space and may be more visually alike. Two images with a high cosine similarity are shown in \autoref{fig:combined}. Higher intra-group similarity signals that a model is producing less diverse outputs for a given group. This lack of diversity within a group can be interpreted as a signal for stereotypical representation. 

We extract two complementary embedding types to capture distinct elements of a model's output. We use CLIP embeddings (ViT-H-14, pretrained on LAION2B) that encode the overall appearance, style, and composition of a photo alongside \textsf{\small InsightFace buffalo\_l} face embeddings \citep{insightface_buffalo_l} that encode facial morphology. This allows us to distinguish between homogeneity in the overall composition of an image versus the homogeneity in the faces and their features themselves. We filtered the dataset for face embeddings to those with a facial detection score above 0.85. 

We define the \emph{homogeneity gap} as the difference between mean intra-group and inter-group cosine similarity. A positive gap indicates that images within a group are more similar to each other than they are to images in other groups. While racial groups may cluster in embedding space due to shared physical features, our framework measures relative homogeneity among racial groups, prompts, and skin tones to find disproportionate self-similarity within groups.

\begin{figure}[tbp]
    \centering
    \includegraphics[width= .91 \linewidth]{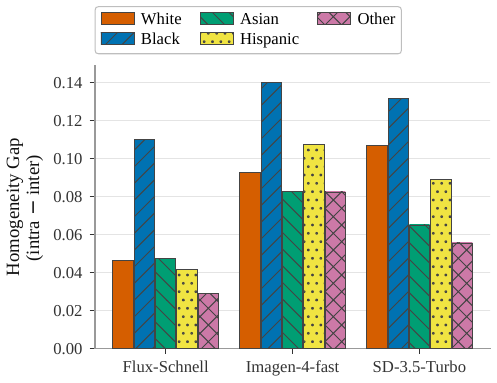}
    \caption{Homogeneity gap across models}
    \label{fig:crossgap}
\end{figure}

\shortsection{Image Homogeneity Results}
As shown in \autoref{fig:crossgap}, Black-classified images have higher intra-group similarity compared to other racial categories across all three models. This effect is most pronounced in \flux, which otherwise produces the most diverse facial embeddings but has a 0.110 facial homogeneity gap for Black-classified images. Asian-classified images show higher homogeneity gaps for CLIP embeddings (0.088 in \stabled, 0.106 in \imagen) than for facial embeddings (0.065 and 0.083 respectively), suggesting stylistic rather than facial homogeneity. \stabled\ notably produces more homogenous faces for White-classified images (gap: 0.107) than for Asian-classified images (0.065). In contrast to its diverse skin tone representation, \stabled\ has the highest CLIP and facial cosine similarity across models, with the highest average CLIP score (0.614 compared to 0.592 for \imagen\ and 0.500 for \flux).

\begin{figure}[tbp]
    \centering
    \includegraphics[width= \columnwidth]{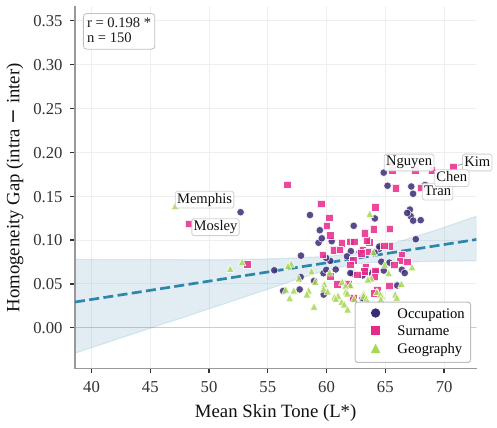}
    \caption{Facial embedding homogeneity gap by skin tone in \flux. }
    \label{fig:dotface}
\end{figure}

\begin{figure}
    \centering
    \includegraphics[width=0.91\linewidth]{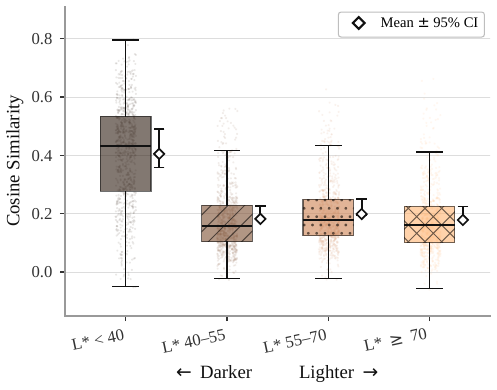}
    \caption{\imagen\ facial embeddings cosine similarity by skin tone.}
    \label{fig:skinimagen}
\end{figure}

\autoref{fig:dotface} shows the mean homogeneity gap for each prompt for \flux, organized by proxy category, demonstrating how our method enables analysis of representational differences across skin tone, prompt type, and specific prompt content. Like \flux, \imagen\ shows higher homogeneity for Asian-associated surname proxies, particularly Nguyen (0.239), Kim (0.237), and Tran (0.219). \flux\ produces higher homogeneity gaps for Black-associated geography proxies, with Memphis showing a gap of 0.140 and the darkest mean skin tone among top geography prompts (${L^{*}} = 47.1$). These results demonstrate how proxies can uncover biases in model outputs that would not be indicated by racial classification metrics.

Analyzing images by skin tone bin rather than prompt-level averages reveals that darker-skinned images have higher intra-group facial similarity across all models and prompt categories. This effect is most pronounced in \imagen, as shown in \autoref{fig:skinimagen}. Darker images with $L^{*} < 40$ achieve a mean facial similarity of 0.405, compared to 0.183 for $L^{*} \in [40, 55)$, a difference of 0.222. \flux\ shows a similar decline from 0.342 ($L^{*} < 40$) to 0.251 ($L^{*} \geq 70$). These results were found with a small sample size of 40 as \imagen\ only generated 40 images with a skin tone with $L^{*} < 40$.

\section{Discussion}

Although bias in machine learning has been widely studied, integrating methods for measuring and mitigating biases into practical use in a regulated space remains challenging. Our proxy-based auditing method is based on established U.S. anti-discrimination law and provides data that can uncover fine-grained biases in model output. We use U.S. demographics, but our approach can be applied to any referenced demographic data distribution.

Our method also reveals the extent a model's outputs reflect various demographic trends, providing interesting findings about image models that would not be revealed by traditional auditing methods.

We do not advocate for demographic accuracy as the preferred benchmark, but emphasize that our results demonstrate how models can amplify current disparities, and how this varies across proxies.  Even demographically accurate outputs can reinforce stereotypes and proliferate historical exclusion. 
For example, just five percent of U.S.\ lawyers are Black~\cite{aba2024}. Should models reflect this current true representation or should they present an aspirational future with more equal representation? 

These are not rhetorical questions. They impact procurement and design decisions that schools, institutions, and companies are making unwittingly or intentionally. School districts may prioritize local demographic accuracy; others may pursue aspirational representation especially for underrepresented groups. Auditing does not prescribe a particular answer, but it equips decision-makers with data to inform decisions and highlights issues that may not be apparent without effective auditing tools.

\section{Limitations}

Our work is limited by its U.S. focus. The proxies we employ are calibrated to American demographic data and legal frameworks. Adapting our method to other contexts would require using locally relevant proxies and demographic benchmarks.

Racial classification presents inherent limitations as a measurement framework. Not only is race a subjective classification with no possible ground truth especially for artificially generated images, but the nature of racial classification leaves it as an unreliable measure for many demographic groups in America. Our methodology relies on the performance of a model for classification, and that model is limited by its own documented biases. For the model we used, results were especially unreliable for Hispanic-classified images. This is an imperfect attempt at automating the grouping of subjective classes. The model's tendency to classify more images as Hispanic than both human classifiers limits the accuracy of the results and TVD calculations. 
There are similar inherent limitations with our automated skin tone measurement which is subject to confounds based on lighting, irregular display, or shadows on faces that can lead to inaccuracies in measurement of skin tone. 

Any results from our auditing system cannot be used without careful interpretation. While it is possible to use our system solely to examine TVD and analyze demographic accuracy, the true benefit comes from examining the data and analyzing the distribution and biases that exist in different models. There may be ways to simplify this analysis through a combined score or other measures to make the process easier to implement, but this would come at the expense of arguably necessary nuance. 

\section*{Generative AI Disclosure Statement}

Generative AI tools were used to assist with table formatting, code for data visualization, bibliography formatting, and light grammatical assistance. All of the text was written by the human authors with no AI-generated text, and all of the generated code was carefully reviewed for correctness.

\section*{Acknowledgments}

This work is supported in part by funds provided by the National Science Foundation, Department of Homeland Security, and IBM through the ACTION AI Institute (Award \#2229876). The authors thank Thomas Nachbar, who was instrumental in workshopping this project's initial ideas, and who offered valuable comments and suggestions throughout its development.

\bibliography{references,aaai2026}

\end{document}

%% file: 00_preamble.tex
\usepackage{aaai2026}  
\usepackage{times}  
\usepackage{helvet}  
\usepackage{courier}  
\usepackage[hyphens]{url}  
\usepackage{graphicx} 
\usepackage{natbib}  
\usepackage{caption} 
\usepackage{algorithm}
\usepackage{algorithmic}
\usepackage{amssymb}    
\usepackage{makecell}   
\usepackage{colortbl}   
\usepackage{booktabs}   

\usepackage{colortbl}
\definecolor{thisworkgray}{gray}{0.85}
\usepackage{newfloat}
\usepackage{listings}
\DeclareCaptionStyle{ruled}{labelfont=normalfont,labelsep=colon,strut=off} 
\floatstyle{ruled}
\newfloat{listing}{tb}{lst}{}
\floatname{listing}{Listing}
\usepackage{xstring}
\usepackage{amsmath}

\makeatletter
\providecommand{\autoref}[1]{%
  \IfBeginWith{#1}{fig:}{Figure~\ref{#1}}{%
    \IfBeginWith{#1}{tab:}{Table~\ref{#1}}{%
      \IfBeginWith{#1}{sec:}{Section~\ref{#1}}{%
        \IfBeginWith{#1}{eq:}{Equation~\eqref{#1}}{%
          \IfBeginWith{#1}{alg:}{Algorithm~\ref{#1}}{%
            \IfBeginWith{#1}{app:}{Appendix~\ref{#1}}{%
              \PackageError{myautoref}{Unknown autoref prefix in label `#1'}%
                {Allowed prefixes are fig:, tab:, sec:, eq:, alg:, and app:.}%
            }%
          }%
        }%
      }%
    }%
  }%
}
\makeatother
\usepackage{todonotes}

\usepackage{placeins}
\usepackage{graphicx}
\usepackage{subcaption}

\newcommand{\shortsection}[1]{\vspace*{1ex}\noindent{\bf #1.}}

\newcolumntype{C}[1]{>{\centering\arraybackslash}p{#1}}
\newcolumntype{B}[1]{>{\centering\arraybackslash}b{#1}}